\documentclass[twoside,11pt]{article}

\begin{document}

\title{Understanding the Meaning of Understanding}

\author{ Daniele Funaro  \\
       Dipartimento di Fisica, informatica e Matematica,\\
       Universita di Modena e Reggio Emilia \\
		daniele.funaro@unimore.it
       }

\date{}
\maketitle

\begin{abstract}
Can we train a machine to detect if another machine has understood a concept?
In principle, this is possible by conducting tests on the subject of that concept. 
However we want this procedure to be done by avoiding direct questions.
In other words, we would like to isolate the absolute meaning of an abstract idea by
putting it into a class of equivalence, hence without adopting straight definitions
or showing how this idea ``works'' in practice.
We discuss the metaphysical implications hidden
in the above question, with the aim of providing a plausible reference framework.
\end{abstract}

\section{Preamble}

\noindent In his famous tale, {\sl Las ruinas circolares} (\cite{borges}), J. L. Borges  depicts the efforts of a man 
in his struggling endeavor to create, through his own dreams, a new life. The accomplishment is to 
induce the dreamed subject to become an independent virtual son, able to live a proper life 
disjointed from the dreams of his creator. The story ends with the stunning discovery that 
the original man is himself the ``materialization" of the dreams of another entity.
\par

The dreamer and the dreamed subject belong to different ``realities", that in the Borges' 
fiction end up to be both ``virtual". At a first glance, there is no direct connection between these universes. 
If ours is somehow the world of reality, the seemingly dissociated domain of our thoughts is usually 
defined as intangible.
One is tempted to attribute  a superior level of abstraction to the second environment, 
though this is not necessarily true, according to the  circularity of Borges' arguments.

\section{Motivations}

\noindent Modern programming instructs machines to learn according to 
apprehending processes that try to mimic those followed by humans. A recurrent question is what an inanimate 
bunch of semiconductors may have understood about the lessons imparted; that is:
how a machine ``visualizes'' in its own ``mind'' the product of new discoveries? Does it
have knowledge of its knowledge? Can other machines know about its thinking ``just by looking
in its eyes''?
These philosophical issues are akin to the thematic of automatic self-consciousness  
(\cite{chris}, \cite{reggia}), a subject that will be rediscussed at the end of section 5.
An attempt to provide some answers is tried in these few pages. It will be argued that
abstract concepts do not follow from definitions or by direct algorithms, but they might be
ruled by the same mechanism that allows to achieve ``understanding'' at the very first level.

\section{Discussion}

\noindent With the supervised help of a teacher, a child can refine the notion of 
color (red, for instance) through examples, by learning how to construct and assign a name to
specific classes of equivalence. Some notions could be actually innate (see, e.g.,  \cite{chomsky}),
however, the individuals of a ``society'', already aware of
those primary concepts, play a fundamental role in the instructing process. By a similar training, a machine 
can recognize if there is a cat on a table. Adding deeper and deeper layers of training, 
the same machine can learn to recognize  a black cat on a wooden table, lapping 
milk from a cup. Despite the increasing complexity of the details, the above training sets 
belong to the space of reality, while the final result (i.e., the knowledge) looks, 
in some way, more ``abstract". The last observation is indeed incorrect from the technical 
viewpoint, since both the images of the cat and the ``neural" outcome of the brain of the instructed 
machine are represented by sequences of the same type of bits. It is only our preconceived 
intuition of reality that tends to assign  different levels to these categories.
\par

At this point, one may ask: how do we know if a child has clear in mind the abstract idea of ``red"? The exam 
is simply done by submitting to his/her attention one or many objects, and pose questions 
about their colors.  Neglecting possible shades of randomness, this analysis is fast and 
secure, since is exactly based on the same apparatus that generates the skill of distinguishing 
colors.
\par

Can we do the above check indirectly, thus without showing any object to the child?  In some parts of the 
child's brain the activation of a certain concept (innate or acquired) has created suitable permanent connections. 
The study of these new links may give an answer to our question without relying on the 
practical experiment. In a very similar way, the machine's concept ``cat on table" resides in a 
memory made of silicon-based circuitry, under the form of a peculiar distribution of data. 
Interpreting these data may teach us if (and maybe what) the machine has formally understood.
\par

Unfortunately, reading a single computer's memory and try to deduce something is like acquiring 
the notion of red through the realization of just one test. Therefore, it is advisable 
to play with a series  of  trained and untrained devices, in order to make comparisons and come 
to conclusions. The path to be followed is the same inspiring the initial training procedure used
for reality. This will be ``supervised" until the learning machine acquires independence. Such an algorithm 
does not necessitate the submission of further cats' pictures. It is an analysis made at a different 
level, like the dream that exists in a more profound layer with respect to that of the dreamer.
The commitment of the learning machine is to distinguish by comparisons the devices that ``know'' from
those that ``do not know'', without studying ``what they actually know''.
It has been already noticed however that all the levels of abstraction are similar from the technical viewpoint.
In truth, following Borges, the dreamer himself is the product of 
the ``imagination" of another dreamer, and, in practice, we have no means to distinguish a {\sl dream} from a {\sl dream into a dream}.

\par

In other words, it would be possible to understand if a device has understood something, 
through a procedure that does not require direct questions. Since this construction is made 
with the help of another training history (at upper level),  we cannot mathematically define (not even a posteriori) 
what kind of configuration must be actually present in a prescribed instructed 
device to be tested. This is not surprising, because it is similar to our incapability of providing an absolute 
definition of red  without indicating an example. ``Red" is a class of equivalence. In the same way, ``have the knowledge
of red" is another class of equivalence; there is no official indication of the elements of this
class, but only examples of elements sharing the same properties. Our mind is modified as we add new
knowledge, however this process is very subtle, so that we cannot practically put into words the details of these changes
(to explain for instance in which area of our brain and under what form those data are present).
\par

The reliability of a set of CPUs, programmed to face AI problems, 
could be in principle verified aseptically, i.e., by plugging electrical supply, but avoiding the use of any
peripherals. 
Here the purpose is not just the check of their plain functionality, but to test their supposed capability 
to apply intelligence. For instance, this kind of training may teach a robot to make a decision not only on the basis of what other robots do, 
but on what they are thinking (if the data of their central systems are available). The instruments to carry out this analysis are standard, although applied to a context
that looks upgraded. As pointed out in a recent article (\cite{rosenfeld}), even current codes for image understanding may fail when tested on 
appropriate nasty examples. What may happen with abstract concepts is at this stage unpredictable, since
it is certainly not an easy task to predict  what is the percentage of trustworthiness 
of these outcomes, which are surely affected by large error spreading. A long phase of experimentation is then necessary. Since this dissertation is only finalized to the
description of the basic principles, we only provide  a few guiding theoretical advices. Thus,
we do not discuss in this paper any concrete development process, leaving to the experts the implementation of the
instances here exposed.

\section{Remarks}

\noindent We proceed in this short disquisition with a warning. Trespassing the privacy of an individual to know if he/she has well 
elaborated the concept of red, without posing straight questions, could be a first step to 
surreptitiously discriminate peoples on the base of
political inclinations, sexual attitudes or whatever an 
organ of control wants to know. It is worthwhile to recall once again that here this kind of analysis is not directly constructed
on specified parameters, but to the belonging or not of the individual to classes
already constituted. The purpose of the machine is to classify  an individual through a characterizing  history
({\sl websites} visited, for example), without examining substantial real facts, but rather
the general activity in the framework of a list of prejudged individuals presenting a well prescribed property.

\par

Without a QI test, a person could be recorded as an ``intelligent fellow" because of his/her affinity to representatives
initially present in that category, with no explicit notion of the properties characterizing that class.
In fact, the setting up of the class itself is the result of a previous analysis carried out
on individuals declared ``intelligent" or ``not intelligent" in advance. Thus, the decision is taken as a consequence
of the ``way to behave", and not on the capability of ``acting intelligently" in the solution of a given problem;
this without the necessity of formalizing officially how an intelligent person is expected to behave.
It is evident that such a superstructure has ethical implications, so that it must be used wisely.

\par

In terms of AI, the memory maps
of a new machine are compared for example with those belonging to the class of ``intelligent machines" and the response is made 
without further external checks. Note that clear traces denoting ``intelligence" could not be present in the preexisting data
(actually, we do not even know how these traces look like),
indeed such files have been just selected on the base of the claimed intelligence of the machines to which they belong.

\par

In this last paragraph the focus is concentrated on another crucial observation.
It is possible to retain the idea of a specific color without naming it. By the way, the
reason why the word ``red'' is the name of the class of all red colored objects,
comes from the necessity of confronting each one's discoveries with other peoples
(see at the beginning of section 3).
The abstraction of the term is actually born in the moment it becomes a product
of the collectivity. Thus,  the interpretation of a thought comes naturally as a
result of the comparison of many minds, as also punctuated before in this paper.

\section{Contextualization}

\noindent The topics touched in this brief exposition are certainly not new 
(\cite{micha}). They assume however
a wider relevance in this specific moment, in which the field of machine learning is 
experiencing a positive period of growth, both in applications and complexity.
In a  recent review paper  (\cite{lecun}), future developments in the field of {\sl deep learning} are
addressed. Among the new disciplines, {\sl reinforcement learning}  is gaining popularity (\cite{sutton}).
There, a progressive tune up of the {\sl policy} is wisely applied to optimize the 
so-called {\sl return}.
For instance, in \cite{singh} and \cite{silver}, this type of training has been implemented without human assisted supervision,
and can represent a first attempt to guide a machine to acquire self-knowledge.
As pointed out several times in this paper, there is a hidden difficulty in going ahead with this construction,
i.e., the device will not be conscious of its own understanding, until this ``state''
is shared with other entities ({\sl I understand we both have understood, because we ``feel'' it
in the same way}).
\par

Efforts have been made in order to
associate increasingly complex concepts, with the help of always more
sophisticated modules acting on data and accomplishing upgraded tasks  (\cite{bottou}). 
In \cite {scassellati}, the concern is to provide a robot with sophisticated skills, 
in order to be able to recognize aspects of the human behavior. 
This implementation, obtained by assembling specialized modules, is a prerogative of a single machine through a process of identification at various stages, similar, 
more or less, to what happens in deep learning. The design requires strong human assistance to be initialized, since 
the building process translates into machine language, the results of the experiences commonly 
lived in reality. The goal is more similar to the effort of creating a sort of human clone, rather than letting the machine to develop a proper way of reasoning.
The above mentioned approaches are then quite different
from the one here discussed, where ``understanding'' is not viewed as a ``complexification'' 
of the bottom, but as a concentrate of the experiences of a community, that can be extracted
on the base of the same principles ruling human connections with reality.  
\par

One may try to establish intersections between our proposal and the so called 
{\sl Theory of Mind} (ToM) (see e.g. \cite{goldman}, \cite{baron}, \cite{wellman}), 
which represents an efficacious instrument of analysis in the sociological and psychological contexts. 
In such a discipline, governed by empiricism, part of the effort is concentrated on the study of the various stages 
of development, where humans acquire knowledge and understanding, through a systematic process named:  
``learning the Theory of Mind''. Again, the translation of these achievements into the machine language seems
to follow a path which is different from what suggested here. In truth, we do not want to teach anything to a computer
or transfer our ``vision'' into it. Instead, the machine has to learn its own ToM. 
For instance, a computer may autonomously build the concept of {\sl wellness}, after 
examining a series of peoples declared by a supervisor to be joyful or sad (maybe because they laugh, cry, or move their face in a bizarre manner). 
The results of this training are, in general, not decipherable, as the machine in its own analysis could emphasize aspects 
of the individuals that we do not even observe or imagine. 
At the end of the process, we do not need to know the definition of ``wellness'' apprehended by the machine. 
On the other hand, if we had a definition of wellness, we could have directly imparted it into the machine from the very beginning.
The learning process is satisfactory if somehow (with a margin of error) the machine has ``understood'' in its own way, and it is able to operate accordingly. There is no need to care about the format of these notions, if the machine can finally do the job, 
for which it has been trained, in the proper way. At higher level, future machines could not necessitate instructions
from men, but they will talk, exchange information, and create new cerebral connections that have nothing in common with those 
usually developed in humans. By following this approach,
in the technology of tomorrow, no human could be in the position to understand what computers actually have in mind.
\par

Going into a more sophisticated area, a possible extension of these considerations can be applied to the field of consciousness,
though the approach may be judged a bit risky (or naive). Consider the phrase:
{\sl I know that I am conscious because I can share this opinion with other peoples, and not because
I can universally define such a feeling}. Again, following this path, the term ``self-knowledge''
applied to an individual turns out to be an element
of a class of equivalence; therefore, it should be studied within this frame of reference.
Thus, based on the material discussed in the present paper, in order to be built, abstraction necessitates of both
``reality'' and the (implicit or explicit) request of a community;
hence it cannot be the consequence of the direct experience of a singlet. 
We can make clear this idea with an example. Let us
suppose that a set of automata learn to play chess and refine their capabilities by continuously challenging each other.
Will be they conscious of being chess players? The answer is {\bf no}, from the simple reason that there is no
utility to develop such a knowledge, unless the machines do ``decide'' together that there is
the necessity to build the class of ``chess players'', with the purpose to distinguish their ability from the state of 
other existing machines that do not even know the basic rules of the game. Recognizing to be part of that class is an 
act of consciousness, although one may argue that this notion is rather {\sl weak} in comparison to more
advanced forms of awareness.  The convenience to give origin to that specific class,
may be due to some (external) forms of gratification. To this purpose, a device may be supplied with {\sl ad hoc} registers
aimed to classify and publicly advertise, the current level of capabilities and a certain degree of ``satisfaction''.
Thus, an isolated single element cannot become conscious by
itself, because, according to our view, such a problem is ill-posed. In contrast to what has been just specified, current research in 
{\sl Artificial Consciousness} is aimed to extract definitions and characterizations in human natural activities 
(\cite{zeman}, \cite{oxford})
to be translated in computational models (see, e.g., \cite{aleksander}, \cite{sun}, \cite{gamez}, as well as
\cite{reggia} for a throughout review of the major achievements). 
Obviously, these approaches follow the reverse path. In \cite{tononi} we can find the following statement:
``consciousness corresponds to the capacity to integrate information". Though we are not moving here in the
direction indicated in that paper, we recognize a vague resemblance with some basic concepts.
\par

\section{Conclusions}

The rationalization process of mathematical type, described so far, involves the classification of objects or abstract 
entities into classes of equivalence. We renounce however to give a definition to the elements of these classes, though we know that each class contains elements 
of the same nature. Classes can be associated with a name (a red hat belongs to both the classes ``red'' and ``hats''). However, names come after the construction 
of a class and are used to communicate to other individuals that something has been apprehended from nature and that
such experiences are waiting to be shared. 
This is different from assuming to have a name (hence, a characterization) and collect together all the entities under that name. 
We cannot create the notion of ``good guy'' from scratch, but we can recognize a good guy among a multitude of fellows. 
This is because our mind, with observations and the exchange of information, has generated the appropriate class of equivalence. 
Classes can be generated at any level of abstraction and complexity. Regarding the viability of these ideas, we are
not able here to investigate further, so that the turn now passes to the experts.
We should however be careful, when establishing parallels between our mind and the work of a machine. Human beings went through a
long process of evolution. Experiences of a single life mix up with innate structures that are inherited from generations, therefore these 
last aspects should not be underestimated.
\par

\vskip 0.2in



\bibliographystyle{plain}

\end{document}